%% file: SCVB0_arXiv.tex
\documentclass[12pt]{article}

\usepackage{amsmath}
\usepackage{amsfonts}
\usepackage{amssymb}
\usepackage{algorithmic}
\usepackage{algorithm}
\usepackage{graphicx}
\usepackage{hyperref}
\usepackage{mdwlist}
\usepackage{multirow}
\usepackage{algorithm}
\usepackage[hmargin=3cm,vmargin=3.5cm]{geometry}
\newtheorem{thm}{Theorem}[section]

\begin{document}

\title{Stochastic Collapsed Variational Bayesian Inference for Latent Dirichlet Allocation}

\author{
James Foulds\textsuperscript{1}\footnote{jfoulds@ics.uci.edu} \and Levi Boyles\textsuperscript{1}\footnote{lboyles@uci.edu} \and Christopher Dubois\textsuperscript{2}\footnote{duboisc@ics.uci.edu} \and Padhraic Smyth\textsuperscript{1}\footnote{smyth@ics.uci.edu}  \and Max Welling\textsuperscript{3}\footnote{m.welling@uva.nl}\\
\textsuperscript{1}Department of Computer Science, University of California, Irvine\\
\textsuperscript{2}Department of Statistics, University of California, Irvine\\
\textsuperscript{3}Informatics Institute, University of Amsterdam\\
}

\maketitle
\begin{abstract}
In the internet era there has been an explosion in the amount of digital text information available, leading to difficulties of scale for traditional inference algorithms for topic models.  Recent advances in stochastic variational inference algorithms for latent Dirichlet allocation (LDA) have made it feasible to learn topic models on large-scale corpora, but these methods do not currently take full advantage of the collapsed representation of the model.  We propose a stochastic algorithm for collapsed variational Bayesian inference for LDA, which is simpler and more efficient than the state of the art method.  We show connections between collapsed variational Bayesian inference and MAP estimation for LDA, and leverage these connections to prove convergence properties of the proposed algorithm.  In experiments on large-scale text corpora, the algorithm was found to converge faster and often to a better solution than the previous method.  Human-subject experiments also demonstrated that the method can learn coherent topics in seconds on small corpora, facilitating the use of topic models in interactive document analysis software.
\end{abstract}

\section{Introduction}

Topic models such as latent Dirichlet allocation (LDA) \cite{blei2003latent} have become a fixture in modern machine learning.  Inference algorithms for topic models provide a low-dimensional representation of text corpora that is typically semantically meaningful, despite being completely unsupervised.  Their use has spread beyond machine learning to become a standard analysis tool for researchers in many fields \cite{mimno2012computational, atkins2012family, mimno2012reconstructing}.  In the internet era, there is a need for tools to learn topic models at the ``web scale'', especially in an industrial setting.  For example, companies such as Yahoo! publish a continually updated stream of online articles, and needs to analyse candidate articles for topical diversity and relevance to current trends, which could be facilitated by topic models.

We would therefore like to have the tools to build topic models that scale to such large corpora, taking advantage of the large amounts of available data to create models that are accurate and contain more topics.  Traditional inference techniques such as Gibbs sampling and variational inference do not readily scale to corpora containing millions of documents or more.  In such cases it is very time-consuming to run even a single iteration of the standard collapsed Gibbs sampling \cite{griffithsSteyvers2004} or variational Bayesian inference algorithms \cite{blei2003latent}, let alone run them until convergence.  The first few passes through the data for these algorithms are inhibited by randomly initialized values that misinform the updates, so multiple such expensive iterations are required to learn the topics.

A significant recent advance was made by Hoffman et al. \cite{hoffman2010online}, who proposed a stochastic variational inference algorithm for LDA topic models.   Because the algorithm does not need to see all of the documents before updating the topics, this method can often learn good topics before a single iteration of the traditional batch inference algorithms would be completed.  The algorithm processes documents in an online fashion, so it can be applied to corpora of any size, or even to never-ending streams of documents.  A more scalable variant of this algorithm was proposed by Mimno et al. \cite{mimno2012sparse}, which approximates the gradient updates in a sparse way in order to improve performance for larger vocabularies and greater numbers of topics.

A complementary direction that has been useful for improving inference in Latent Dirichlet allocation is to take advantage of its ``collapsed'' representation, where parameters are marginalized out, leaving only latent variables.  It is possible to perform inference in the collapsed space and recover estimates of the parameters afterwards.  For existing inference techniques that operate in a batch setting, the algorithms that operate in the collapsed space are more efficient at improving held-out log probability than their uncollapsed counterparts, both per iteration and in wall-clock time per iteration \cite{griffithsSteyvers2004, teh2007collapsed, asuncionsmoothing}.  Reasons for this advantage include the per-token updates which propagate updated information sooner, simpler update equations, fewer parameters to update, no expensive calls to the digamma function, and the avoidance of tightly coupled pairs of parameters which inhibit mixing for Gibbs sampling \cite{carpenter2010collapsed, asuncionsmoothing, teh2007collapsed}.  For variational inference, perhaps the most important advantage of the collapsed representation is that the variational bound is strictly better than for the uncollapsed representation, leading to the potential to learn more accurate topic models \cite{teh2007collapsed}.  The existing online inference algorithms for LDA do not fully take advantage of the collapsed representation -- although the sparse online LDA algorithm of Mimno et al. \cite{mimno2012sparse} collapses out per-document parameters $\theta$, the topics themselves are not collapsed so there is no improvement in the variational bound.

In this work, we develop a stochastic algorithm for LDA that operates fully in the collapsed space, thus transferring the aforementioned advantages of collapsed inference to the online setting.  This facilitates learning topic models both more accurately and more quickly on large datasets.  The proposed algorithm is also very simple to implement, requiring only basic arithmetic operations.  We show that from another perspective, the algorithm can also be interpreted as a MAP estimation algorithm.  This interpretation allows us to prove the convergence of the algorithm.  We also explore the benefit of our method on small problems, showing that it is feasible to learn human-interpretable topics in seconds.

\section{Background}

Probabilistic topic models such as Latent Dirichlet allocation (LDA) \cite{blei2003latent} use latent variables to encode co-occurrence patterns between words in text corpora, and other bag-of-words data.  In the LDA model, there are $K$ topics $\phi_k, k \in \{1, \ldots, K\}$, which are discrete distributions over words.  For example, a topic on baseball might give high probabilities to words such as ``pitcher'', ``bat'' and ``base''.  The assumed generative process for the LDA model is 

\begin{algorithmic}
  \STATE Generate each topic $\phi_k \sim \mbox{Dirichlet}(\eta), k \in \{1, \ldots, K\}$ \\
  \STATE For each document $j$\\
  \STATE \hspace{\algorithmicindent} Generate a distribution over topics $\theta_j \sim \mbox{Dirichlet} (\alpha$) \\
  \STATE \hspace{\algorithmicindent} For each word $i$ in document $j$\\
  \STATE \hspace{\algorithmicindent} \hspace{\algorithmicindent} Sample a topic $z_{ij} \sim \mbox{Discrete}(\theta_j)$\\
  \STATE\hspace{\algorithmicindent} \hspace{\algorithmicindent} Sample the word $w_{ij} \sim \mbox{Discrete}(\phi_{z_{ij}})$ \mbox{  .}\\
\end{algorithmic}


To scale LDA inference to very large datasets, a stochastic variational inference algorithm was proposed by Hoffman et al. \cite{hoffman2010online}.  We will discuss its more general form \cite{hoffman2012stochastic}, which applies to all graphical models whose parameters can be split into ``global'' parameters $G$ and ``local'' parameters $L_j$ pertaining to each data point $x_j$, with complete conditionals being exponential family distributions.  The algorithm examines one data point at a time to learn that data point's ``local'' variational parameters, such as $\theta_j$ in LDA.  It then updates ``global'' variational parameters, such as topics $\phi_k$, via a stochastic natural gradient update.  Their general scheme is given in Algorithm \ref{alg:hoffman}.

\begin{algorithm}
\caption{Stochastic Variational Inference (Hoffman et al.) \label{alg:hoffman}}
\begin{itemize*}
\item Input: Data $x_1, \dots, x_D$, step sizes $\rho_t$, $t=1:M$ (Max iterations)
\item Randomly initialize ``global'' (e.g. topic) parameters $G$
\item  For $t = 1:M$
	 \begin{itemize*}
	     \item Select a random data point (e.g. document) $x_j$, $j \in \{1, \ldots, D\}$
  			\item Compute ``local'' (e.g. document-level) variational parameters $\mathbf{L}_j$
				\item $\hat{\mathbf{G}} = D\mathbf{L}_j$
				\item $\mathbf{G} := (1 - \rho_t) \mathbf{G} + \rho_t \hat{\mathbf{G}}$
	 \end{itemize*}
\end{itemize*}
\end{algorithm}

For an appropriate local update and sequence of step sizes $\rho$, this algorithm is guaranteed to converge to the optimal variational solution \cite{hoffman2012stochastic}.  In the case of LDA, let $\mathbf{\lambda}_k$ be the parameter vector for a variational Dirichlet distribution on topic $\phi_k$.  This method computes variational distributions for topic assignments and the distribution over topics for document $j$ using a regular VB update, then for each topic $k$ computes $\hat{\lambda_k}$, an estimate for what $\lambda_k$ would be if all $D$ documents were identical to document $j$.  It then updates the $\lambda_k$'s via a natural gradient update, which takes the form

\begin{equation}
\label{eqn:hoffman}
\lambda_k := (1 - \rho_t)\lambda_k + \rho_t \hat{\lambda}_k \mbox{ .}
\end{equation}

The online EM algorithm of Cappe and Moulines \cite{cappe2009line} is another general-purpose method for learning latent variable models in an online setting.  The algorithm alternates between an ordinary M-step which maximizes the EM lower bound with respect to parameters $\theta$, and a stochastic expectation step, which updates exponential family sufficient statistics $\mathbf{s}$ with an online average

\begin{equation}
\label{eqn:onlineEM}
\mathbf{s} := (1 - \rho_t)s + \rho_t \hat{\mathbf{s}}(Y_{n+1};\theta) \mbox{ ,}
\end{equation}

with $Y_{n+1}$ being a new data point, $\theta$ being the current parameters, and $\hat{\mathbf{s}}(Y_{n+1};\theta)$ being an estimate of the sufficient statistics based on these values.

In this article, we show how to perform stochastic variational inference in the collapsed representation of LDA, using an algorithm inspired by the online algorithms of Hoffman et al. and Cappe and Moulines.  The new algorithm takes advantage of a fast collapsed inference method called ``CVB0'' \cite{asuncionsmoothing} to further improve the efficiency of stochastic LDA inference.

\subsection{CVB0}
In the collapsed representation of LDA, we marginalize out topics $\mathbf{\Theta}$ and distributions over topics $\mathbf{\Phi}$, and perform inference only on the topic assignments $\mathbf{Z}$.  The collapsed variational Bayesian inference (CVB) approach of Teh et al. \cite{teh2007collapsed} maintains variational discrete distributions $\gamma_{ij}$ over the $K$ topic assignment probabilities for each word $i$ in each document $j$.   Teh et al. showed that although the updates for a coordinate ascent algorithm optimizing the evidence lower bound with respect to $\gamma$ are intractable, an algorithm using approximate updates works well in practice, outperforming the classical VB algorithm in terms of prediction performance.  Asuncion et al. \cite{asuncionsmoothing} showed that a simpler version of this method, called CVB0, is much faster while still maintaining the accuracy of CVB.  The CVB0 algorithm iteratively updates each $\gamma_{ij}$ via

\begin{equation}
 \label{eqn:cvb0}
  \gamma_{ij k} :\propto \frac{N_{w_{ij}k}^{\Phi\neg ij} + \eta_{w_{ij}}}{N_k^{Z \neg ij} + \sum_w \eta_w} (N_{jk}^{\Theta\neg ij} + \alpha)  \mbox{ ,}
\end{equation}

for each topic $k$, with $w_{ij}$ corresponding to the word token for the $j$th document's $i$th word. The $\mathbf{N}^Z$, $\mathbf{N}^\Theta$ and $\mathbf{N}^\Phi$ variables, henceforth referred to as the ``CVB0 statistics'', are variational expected counts corresponding to their indices, and the $\neg ij$ superscript indicates the exclusion of the current value of $\gamma_{ij}$.    Specifically, $\mathbf{N}^Z$ is the vector of expected number of words assigned to each topic, $\mathbf{N}^\Theta_{j}$ is the equivalent vector for document $j$ only, and each entry $w,k$ of matrix $\mathbf{N}^\Phi$ is the expected number of times word $w$ is assigned to topic $k$ across the corpus,
\begin{align}
 N^Z_k &\triangleq \sum_{ij} \gamma_{ijk} &&&
 N^{\Theta}_{jk} &\triangleq  \sum_i \gamma_{ijk} &&&
 N^{\Phi}_{w k} &\triangleq  \sum_{ij: w_{ij} = w} \gamma_{ij k} \mbox{ .}
\end{align}

Note that $\mathbf{N}^\Theta_{j} + \alpha$ is an unnormalized variational estimate of the posterior mean of document $j$'s distribution over topics $\mathbf{\theta}_j$, and column $k$ of $\mathbf{N}^\Phi + \beta$ is an unnormalized variational estimate of the posterior mean of topic $\mathbf{\phi}_k$.  The update for CVB0 closely resembles the collapsed Gibbs update for LDA, but is deterministic.

CVB0 is currently the fastest technique for LDA inference for single-core batch inference in terms of convergence rate \cite{asuncionsmoothing}.   It is also as simple to implement as collapsed Gibbs sampling, and has a very similar update procedure.  Sato and Nakagawa \cite{sato2012rethinking} showed that the terms in the CVB0 update can be understood as optimizing the $\alpha$-divergence, with different values of $\alpha$ for each term.  The $\alpha$-divergence is a generalization of the KL-divergence that variational Bayes minimizes, and optimizing it is known as power EP \cite{minka2004power}.  A disadvantage of CVB0 is that the memory requirements are large as it needs to store a variational distribution $\gamma$ for every token in the corpus.  This can be improved slightly by ``clumping'' every occurrence of a specific word in each document together and storing a single $\gamma$ for them.

\section{Stochastic CVB0}

We would like to exploit the efficiency and simplicity of CVB0, and the improved variational bound of the collapsed representation in a stochastic algorithm.  Such an algorithm should not need to maintain the $\gamma$ variables, thus circumventing the memory requirements of CVB0, and should be able to provide an estimate for the topics when only a subset of the data have been visited.  Recall that the CVB0 statistics $\mathbf{N}^Z$, $\mathbf{N}^\Theta$ and $\mathbf{N}^\Phi$ are all that are needed to both perform a CVB0 update and to recover estimates of the topics.  So, we want to be able to estimate the CVB0 statistics based on the set of tokens we have observed.  

Suppose we have seen a token $w_{ij}$, and its associated $\mathbf{\gamma}_{ij}$.  The information this gives us about the statistics depends on how the token was drawn.  If the token was drawn uniformly at random from all of the tokens in the corpus, the expected value of $\mathbf{N}^Z$ with respect to the sampling distribution is $C\mathbf{\gamma}_{ij}$, where $C$ is the number of words in the corpus.  For the same sampling procedure, the expectation of the word-topic expected counts matrix $\mathbf{N}^\Phi$ is $C\mathbf{Y}^{(ij)}$, where $\mathbf{Y}^{(ij)}$ is a $W \times K$ matrix with the $w_{ij}$th row being $\mathbf{\gamma}_{ij}$ and with zeros in the other entries.  Now if the token was drawn uniformly from the tokens in document $j$, the expected value of $\mathbf{N}^\Theta_j$ is $C_{j}\mathbf{\gamma}_{ij}$.\footnote{Other sampling schemes are possible, which would lead to different algorithms.  For example, one could sample from the set of tokens with word index $w$ to estimate $\mathbf{N}^\Phi_{w}$.  Our choice leads to an algorithm that is practical in the online setting.}

Since we may not maintain the $\gamma$'s, we cannot perform these sampling procedures directly.  However, with a current guess at the CVB0 statistics we can \emph{update} a token's variational distribution, and observe its new value.  We can then use this $\gamma_{ij}$ to improve our estimate of the CVB0 statistics.  This suggests an iterative procedure, alternating between a ``maximization'' step, approximately optimizing the evidence lower bound with respect to a particular $\gamma_{ij}$ via CVB0, and an ``expectation'' step, where we update the expected count statistics to take into account the new $\gamma_{ij}$.  As the algorithm continues, the $\gamma_{ij}$'s we observe will change, so we cannot simply average them.  Instead, we can follow Cappe and Moulines \cite{cappe2009line} and perform an online average of these statistics via Equation \ref{eqn:onlineEM}.

In the proposed algorithm, we process the corpus one token at a time, examining the tokens from each document in turn.  For each token, we first compute a new $\gamma_{ij}$.  We do not store the $\gamma$'s, but compute (updated versions of) them as needed via CVB0.  This means we must make a small additional approximation in that we cannot subtract current values of $\gamma_{ij}$ in Equation \ref{eqn:cvb0}.  With large corpora and large documents this difference is negligible.  The update becomes

\begin{equation}
 \label{eqn:cvb0_nocount}
  \gamma_{ij k} :\propto \frac{N_{w_{ij} k}^{\Phi} + \eta_{w_{ij}}}{N^Z_k + \sum_w \eta_w} (N_{jk}^{\Theta} + \alpha) \mbox{ .}
\end{equation}

We then use this to re-estimate our CVB0 statistics.  We use one sequence of step-sizes $\rho^\Phi$ for $\mathbf{N}^\Phi$ and $\mathbf{N}^Z$, and another sequence $\rho^\Theta$ for $\mathbf{N}^\Theta$.  While we are processing randomly ordered tokens $i$ of document $j$, we are effectively drawing random tokens from it, so the expectation of $\mathbf{N}^\Theta_{j}$ is $C_{j}\gamma_{ij}$. We update $\mathbf{N}^\Theta_{j}$ with an online average of the current value and its expected value,
\begin{equation}
\label{eqn:NTheta}
\mathbf{N}^\Theta_{j} := (1 - \rho^{\Theta}_{t})\mathbf{N}^\Theta_{j}  + \rho^{\Theta}_{t} C_{j}\gamma_{ij} \mbox{ .}
\end{equation}

Although we process a document at a time, we eventually process all of the words in the corpus.  So for the purposes of updating $\mathbf{N}^\Phi$ and $\mathbf{N}^Z$, in the long-run we are effectively drawing tokens from the entire corpus.  The expected $\mathbf{N}^\Phi$ after observing one $\gamma_{ij}$ is $C\mathbf{Y}^{(ij)}$, and the expected $\mathbf{N}^Z$ is $C\gamma_{ij}$.   In practice, it is too expensive to update the entire $\mathbf{N}^\Phi$ after every token, suggesting the use of minibatch updates.  The expected $\mathbf{N}^\Phi$ after observing a minibatch $M$ is the average of the per-token estimates, and similarly for $\mathbf{N}^Z$, leading to the updates:

\begin{align}
  \mathbf{N}^\Phi &:= (1 - \rho^{\Phi}_{t})\mathbf{N}^\Phi  + \rho^{\Phi}_{t}\hat{\mathbf{N}}^\Phi  \label{eqn:NPhi}\\
  \mathbf{N}^Z &:= (1 - \rho^{\Phi}_{t})\mathbf{N}^Z + \rho^{\Phi}_{t} \hat{\mathbf{N}}^Z \label{eqn:Nbullet}
\end{align}

where $\hat{\mathbf{N}}^\Phi = \frac{C}{|M|}   \sum_{ij \in M} \mathbf{Y}^{(ij)}$ and $\hat{\mathbf{N}}^Z = \frac{C}{|M|} \sum_{ij \in M} \gamma_{ij}$.
Depending on the lengths of the documents and the number of topics, it is often also beneficial to perform a small number of extra passes to learn the document statistics before updating the topic statistics.  We found that one such burn-in pass was sufficient in all of the datasets we tried in our experiments.  Pseudo-code for the algorithm, which we refer to as ``Stochastic CVB0'' (SCVB0) is given in Algorithm \ref{alg:scvb0}.

\begin{algorithm}
\caption{\label{alg:scvb0} Stochastic CVB0 \label{alg:scvbperTokenMinibatch}}
\begin{itemize*}
\item Randomly initialize $\mathbf{N}^\Phi$, $\mathbf{N}^\Theta$; $\mathbf{N}^Z :=\sum_w \mathbf{N}^\Phi_w$
\item $\hat{\mathbf{N}}^\Phi := \mathbf{0}$; $\hat{\mathbf{N}}_\bullet  := \mathbf{0}$
\item  For each document $j$
  \begin{itemize*}
		 \item For zero or more ``burn-in'' passes
		 \begin{itemize*}
		     \item For each token $i$
		     \begin{itemize*}
			 		   \item Update $\gamma_{ij}$ \mbox{  }(Equation \ref{eqn:cvb0_nocount})
					   \item Update $\mathbf{N}^\Theta_{j}$ \mbox{  }(Equation \ref{eqn:NTheta})
					\end{itemize*}
		 \end{itemize*}
		 
		 \item For each token $i$
		 \begin{itemize*}
			 		\item Update $\gamma_{ij}$ \mbox{  }(Equation \ref{eqn:cvb0_nocount})
					\item Update $\mathbf{N}^\Theta_{j}$ \mbox{  }(Equation \ref{eqn:NTheta})
					\item $\hat{\mathbf{N}}_{w_t} := \hat{\mathbf{N}}_{w_t} + C\gamma_{ij}$
				  \item $\hat{\mathbf{N}}^Z := \hat{\mathbf{N}}^Z + C\gamma_{ij}$
		 \end{itemize*}
		\item If minibatch finished
\begin{itemize*}
						 \item Update $\mathbf{N}^\Phi$ \mbox{  }(Equation \ref{eqn:NPhi})
						 \item Update $\mathbf{N}^Z$ \mbox{  }(Equation \ref{eqn:Nbullet})
						 \item $\hat{\mathbf{N}}^\Phi := \mathbf{0}$; $\hat{\mathbf{N}}^Z  := \mathbf{0}$
		\end{itemize*}
	\end{itemize*}
	\end{itemize*}
\end{algorithm}

An optional optimization to the above algorithm is to only perform one update for each distinct token in each document, and scale the update by the number of copies in the document.  This process, often called ``clumping'', is standard practice for fast implementations of all LDA inference algorithms, though it is only exact for uncollapsed algorithms, where the $z_{ij}$'s are D-separated by $\theta_j$.  Suppose we have observed $w_{tj}$, which occurs $m_{tj}$ times in document $j$.  Plugging Equation \ref{eqn:NTheta} into itself $m_{tj}$ times and noticing that all but one of the resulting terms form a geometric series, we can see that performing $m_{tj}$ updates for $\mathbf{N}^\Theta_{j}$ while holding $\gamma_{ij}$ fixed is equivalent to

\begin{equation}
\mathbf{N}^\Theta_{j} := (1-\rho^\Theta_t)^{m_{tj}}\mathbf{N}^\Theta_{j} + C_j\gamma_{ij} (1 - (1 - \rho^\Theta_t)^{m_{tj}}) \mbox{  .}
\end{equation}

\section{An Alternative Perspective: MAP Estimation}

 In the SCVB0 algorithm, because the $\gamma$'s are not maintained we must approximate Equation \ref{eqn:cvb0} with Equation \ref{eqn:cvb0_nocount}, neglecting the subtraction of the previous value of $\gamma_{ij}$ from the CVB0 statistics when updating $\gamma_{ij}$.  It can be shown that this approximation results in an algorithm which is equivalent to an EM algorithm for MAP estimation, due to Asuncion et al. \cite{asuncionsmoothing}, which operates on an unnormalized parameterization of LDA.  Therefore, the approximate collapsed variational updates of SCVB0 can also be understood as MAP estimation updates.  Using this interpretation of the algorithm, we now give an alternative derivation of SCVB0 as a version of Cappe and Moulines' online EM algorithm \cite{cappe2009line} as applied to MAP estimation for LDA, thus providing an alternative perspective on the algorithm.

In particular, iterating the following update optimizes an EM lower bound on the posterior probability of the parameters:

\begin{equation}
 \label{eqn:map_lda}
  \bar{\gamma}_{ij k} :\propto \frac{\bar{N}^{\Phi}_{w_{ij} k} + \eta - 1}{\bar{N}^Z_k + W (\eta - 1)} (\bar{N}^{\Theta}_{jk} + \alpha - 1) \mbox{ ,}
\end{equation}

where $\bar{\gamma}_{ijk} \triangleq Pr(z_{ij} = k| \bar{N}^{\Phi}, \bar{N}^Z, \bar{N}^{\Theta}, w_{ij})$ are EM ``responsibilities'', and the other variables, which we will refer to as \emph{EM statistics}, are aggregate statistics computed from sums of these responsibilities,

\begin{align}
 \label{eqn:emstat1} \bar{N}^Z_k =\sum_{ij} \bar{\gamma}_{ijk} &&
 \bar{N}^{\Theta}_{jk} =  \sum_i \bar{\gamma}_{ijk} &&
  \bar{N}^{\Phi}_{w k} =  \sum_{ij: w_{ij} = w} \bar{\gamma}_{ij k} \mbox{ .} 
\end{align}

Upon completion of the algorithm, MAP estimates of the parameters can be recovered by 

\begin{align}
\label{eqn:recoverMAP}
 \hat{\phi}_{wk} = \frac{\bar{N}^{\Phi}_{wk} + \eta - 1}{\bar{N}^Z_k + W (\eta - 1)} && \hat{\theta}_{jk} = \frac{\bar{N}^{\Theta}_{jk} + \alpha - 1}{C_j + K\alpha - K} \mbox{ ,}
\end{align}

where $C_j$ is the length of document $j$.  A sketch of the derivation for this algorithm, which we will refer to as \emph{unnormalized MAP LDA} (MAP\_LDA\_U), is given in Appendix \ref{sec:app_unnormalizedMAP}.  Note that if we identify the EM statistics and responsibilities with CVB0 statistics and variational distributions, the SCVB0 update in Equation \ref{eqn:cvb0_nocount} is identical to Equation \ref{eqn:map_lda} but with the hyper-parameters adjusted by one.  

We now adapt online EM to this setting.  In its general form, the online EM algorithm performs maximum likelihood estimation by alternating between updating an online estimate of the expected sufficient statistics for the complete-data log-likelihood and optimizing parameter estimates via a regular EM M-step.  We consider this algorithm as applied to an unnormalized parameterization of LDA, where the parameters of interest are estimates $\mathbf{\hat{\bar{N}}^{\Phi}}$, $\mathbf{\hat{\bar{N}}^{\Theta}}$, $\mathbf{\hat{\bar{N}}^{Z}}$ of the EM statistics, which are related to $\Theta$ and $\Phi$ via Equation \ref{eqn:recoverMAP}.  We also adapt the algorithm to perform MAP estimation, and to operate with stochasticity at the word-level instead of the document-level.  The resulting algorithm is procedurally identical to SCVB0.

First, let us derive the expected sufficient statistics.  Written in exponential family form, the complete data likelihood for a word $w_{ij}$ and its topic assignment $z_{ij}$ is

\begin{align}
 \nonumber
\exp \Big(&\sum_{wk} [w_{ij} = w][z_{ij} = k]\log( \frac{\hat{\bar{N}}^{\Phi}_{w k} + \eta - 1}{\hat{\bar{N}}^Z_k + W (\eta - 1)})\\ \nonumber
                & + \sum_{k} [z_{ij} = k]\log (\frac{\hat{\bar{N}}^{\Theta}_{jk} + \alpha - 1}{N_j + K(\alpha - 1)}) \Big)\\ \nonumber
\propto \exp \Big(&\sum_{wk} [w_{ij} = w][z_{ij} = k]\log(\hat{\bar{N}}^{\Phi}_{w k} + \eta - 1)\\ \nonumber
+&\sum_{k} [z_{ij} = k]\log (\hat{\bar{N}}^{\Theta}_{jk} + \alpha - 1)\\
-&\sum_{k} [z_{ij} = k]\log(\hat{\bar{N}}^Z_k + W (\eta - 1)) \Big) \mbox{ ,}
\end{align}

where $[a = b]$ is a Kronecker delta function, equal to one if $a = b$ and zero otherwise, and $\hat{\bar{N}}$ variables denote current estimates, not necessarily synchronized with $\bar{\gamma}$.  We can see that the sufficient statistics are the delta functions (and products of delta functions),

\begin{align}
 \nonumber
 S^{(w)}(w_{ij}, z_{ij}) =  (&[w_{ij} = 1][z_{ij} = 1], \ldots, [w_{ij} = W][z_{ij} = K], \\
  &[z_{ij} = 1], \ldots, [z_{ij} = K], [z_{ij} = 1], \ldots, [z_{ij} = K])^\intercal \mbox{ ,}
\end{align}

 and the expected sufficient statistics, given current parameter estimates, are appropriate entries of $\bar{\gamma}$,

\begin{align}
 \nonumber
\bar{s}^{(w)}(w_{ij}, z_{ij}) =  (&[w_{ij} = 1]\bar{\gamma}_{ij1}, \ldots, [w_{ij} = W]\bar{\gamma}_{ijK}, \\
  &\bar{\gamma}_{ij1}, \ldots, \bar{\gamma}_{ijK}, \bar{\gamma}_{ij1}, \ldots, \bar{\gamma}_{ijK})^\intercal \mbox{ .}
\end{align}

Cappe and Moulines normalize the likelihood, and the sufficient statistics, by the number of data points $n$, so that $n$ need not be specified in advance.  However, since we are performing MAP estimation, unlike the MLE algorithm described by Cappe and Moulines, we need to estimate the unnormalized expected sufficient statistics for the entire corpus in order to maintain the correct scale relative to the prior.  This can be achieved by scaling the per-word expected sufficient statistics by appropriate constants to match the size of the corpus (or document, for per-document statistics)

\begin{align}
\label{eqn:scaledESS} \nonumber
\bar{s}^{\prime(w)}(w_{ij}, z_{ij}) =  (&C[w_{ij} = 1]\bar{\gamma}_{ij1}, \ldots, C[w_{ij} = W]\bar{\gamma}_{ijK}, \\
  &C_j\bar{\gamma}_{ij1}, \ldots, C_j\bar{\gamma}_{ijK}, C\bar{\gamma}_{ij1}, \ldots, C\bar{\gamma}_{ijK})^\intercal \mbox{ .}
\end{align}

Notice that the average of these corpus-wide expected sufficient statistics, computed across all tokens in the corpus, is equal to the EM statistics,  i.e. the parameters to be optimized in the M-step.  Collecting them into appropriate matrices, we can write the the expected sufficient statistics as

\begin{align}
 \hat{\bar{s}} = (\mathbf{\bar{N}^\Theta}, \mathbf{\bar{N}^\Phi}, \mathbf{\bar{N}^Z}) \mbox{ .}
\end{align}

In fact, optimizing the EM objective function with respect to the parameters, we find that the M-step assigns the parameter estimate EM statistics to be consistent with the EM statistics computed in the E-step (CF Appendix \ref{sec:app_unnormalizedMAP}),
\begin{align}
\label{eqn:unnormalized_map_m_step}
\mathbf{\hat{\bar{N}}^\Theta} &:= \mathbf{\bar{N}^\Theta} &&&
\mathbf{\hat{\bar{N}}^\Phi} &:= \mathbf{\bar{N}^\Phi} &&&
\mathbf{\hat{\bar{N}}^Z} &:= \mathbf{\bar{N}^Z} \mbox{ .}
\end{align}

We therefore do not need to store parameter estimates $\mathbf{\hat{\bar{N}}}$ separately from expected sufficient statistics $\mathbf{\bar{N}}$, as M-step updated parameter estimates are always equal to the expected sufficient statistics from the E-step.  Inserting Equation \ref{eqn:scaledESS} into Equation \ref{eqn:onlineEM} and using separate step size schedules for document statistics and topic statistics, the online E-step after processing token $w_{ij}$ is given by

\begin{align}
\mathbf{\bar{N}}^\Theta_{j} &:= (1 - \rho^{\Theta}_{t})\mathbf{\bar{N}}^\Theta_{j}  + \rho^{\Theta}_{t} C_{j}\gamma_{ij} \\
\mathbf{\bar{N}}^\Phi_w &:= (1 - \rho^{\Phi}_{t})\mathbf{\bar{N}}^\Phi  + \rho^{\Phi}_{t}C\gamma_{ij}[w_{ij} = w] \mbox{ , } \forall w \\
  \mathbf{\bar{N}}^Z &:= (1 - \rho^{\Phi}_{t})\mathbf{\bar{N}}^Z + \rho^{\Phi}_{t} C\gamma_{ij} \mbox{ ,}
\end{align}

with $\bar{\gamma}_{ij}$ computed via Equation \ref{eqn:map_lda}.  The online EM algorithm we have just derived is procedurally identical to SCVB0 with minibatches of size one, identifying EM responsibilities and statistics with SCVB0 responsibilities and statistics, and with the hyper-parameters adjusted by one.  Under this interpretation, an alternative name for SCVB0 might be \emph{stochastic unnormalized MAP LDA} (S\_MAP\_LDA\_U).

\section{Convergence Analysis}
\label{sec:convergence}
The MAP estimation interpretation of SCVB0 is the interpretation that is most amenable to convergence analysis, since MAP\_LDA\_U exactly optimizes a well-defined objective function, while CVB0 has approximate updates.  In this section, the notation will follow the MAP interpretation of the algorithm.  We have the following theorem:

\begin{thm}
If $0 < \rho_t^\Phi \leq 1 \mbox{ } \forall t$, $0 < \rho_t^\Theta \leq 1 \mbox{ } \forall t$, $\sum_{t=1}^\infty \rho_t^\Phi = \infty$, $\lim_{t \rightarrow \infty} \rho_t^{\Phi} = 0$, $\sum_{t=1}^\infty \rho_t^\Theta = \infty$, and $\lim_{t \rightarrow \infty} \rho_t^{\Theta } = 0$, then in the limit as the number of iterations $t$ approaches infinity SCVB0 converges to a stationary point of the MAP objective function.
\end{thm}
\emph{Proof} Consider the MAP\_LDA\_U algorithm, with an update schedule alternating between a full E-step, i.e. updating every $\bar{\gamma}_{ij}$, and a full M-step, i.e. synchronizing the (parameter estimate) EM statistics with the $\bar{\gamma}$'s.  The $\bar{\gamma}$'s do not depend on each other given the EM statistics, so we do not need to maintain them between iterations.  We can thus view this version of MAP\_LDA\_U as operating on just the EM statistics.  For each EM statistic $c \in \{ \mathbf{\hat{\bar{N}}^\Theta_1}, \ldots, \mathbf{\hat{\bar{N}}^\Theta_D}, \mathbf{\hat{\bar{N}}^\Phi}, \mathbf{\hat{\bar{N}}^Z} \}$, let $f_c(X, \hat{s}): S_c \rightarrow S_c$ be a mapping from a current value to the updated value after such an iteration, i.e. performing an E-step to estimate the $\bar{\gamma}$'s, then using these to update the parameter estimates in the M-step, where $X$ is the full corpus and $S_c$ is the space of possible assignments for EM statistic $c$.

Let $\hat{\bar{s}} = (\mathbf{\hat{\bar{N}}^\Theta_1}, \ldots, \mathbf{\hat{\bar{N}}^\Theta_D}, \mathbf{\hat{\bar{N}}^\Phi}, \mathbf{\hat{\bar{N}}^Z})$ be an assignment of the EM statistics, with $\hat{\bar{s}}_c$ referring to EM statistic $c$, and let $\hat{\bar{s}}^{(t)}$ be the EM statistics at word iteration $t$ of the SCVB0 algorithm.  Furthermore, let $\bar{s}_c(w^{(t+1)}, \hat{s})$ be the estimate of $f_c(X, \hat{s})$ based on the word $w^{(t+1)}$ examined at step $t+1$, as per the right hand side of the SCVB0 update equations.   Note that $E[\bar{s}_c(w^{(t+1)}, \hat{s})] = f_c(X, \hat{s})$, where the expectation is with respect to the sampling of $w^{(t+1)}$.  Finally, let $\xi^{(t+1)} = \bar{s}_c(w^{(t+1)}, \hat{s}^{(t)}) - f_c(X, \hat{s}^{(t)})$ be the stochastic error made at step $t+1$, and observe that $E[\xi^{(t+1)}] = 0$.  We can rewrite the SCVB0 updates for each EM statistic $c$ as 

\begin{align}
 \nonumber
 \hat{\bar{s}}^{(t+1)}_c &= (1- \rho^c_{t+1})\hat{\bar{s}}^{(t)}_c + \rho^c_{t+1}\bar{s}_c(w^{(t+1)}, \hat{\bar{s}})\\  \nonumber
   &= \hat{\bar{s}}^{(t)}_c + \rho^c_{t+1}( -\hat{\bar{s}}^{(t)}_c + \bar{s}_c(w^{(t+1)}, \hat{\bar{s}}))\\  \nonumber
   &= \hat{\bar{s}}^{(t)}_c + \rho^c_{t+1}(f_c(X, \hat{\bar{s}}^{(t)}) -\hat{\bar{s}}^{(t)}_c + \bar{s}_c(w^{(t+1)}, \hat{\bar{s}}) - f_c(X, \hat{\bar{s}}^{(t)}))\\
   &= \hat{\bar{s}}^{(t)}_c + \rho^c_{t+1}(f_c(X, \hat{\bar{s}}^{(t)}) -\hat{\bar{s}}^{(t)}_c + \xi^{(t+1)}) \mbox{ .}
\end{align}

In this form, we can see that iterating each of the SCVB0 updates corresponds to a Robbins-Monro stochastic approximation (SA) algorithm \cite{robbins1951stochastic} for finding the zeros of $f_c(X, \hat{s}^{(t)}) -\hat{s}^{(t)}_c$, i.e. the fixed points of MAP\_LDA\_U for $\hat{s}_c$.  Since MAP\_LDA\_U is an EM algorithm, its fixed points are the stationary points of the posterior probability of the parameters, as recovered via Equation \ref{eqn:recoverMAP}.

Theorem 2.3 of Andreiu et al. \cite{andrieu2005stability} states that under mild conditions, the existence of a Lyapunov function, along with a boundedness condition, implies that such a Robins-Monro algorithm will converge with step size schedules such as those above.  In the context of an SA algorithm, a Lyapunov function can be understood as an ``objective function'' which, in the absence of stochastic noise, the SA would improve monotonically if small enough steps were taken in the direction of the updates.  In Appendix \ref{sec:app_lyapunov}, we show that the negative of the Lagrangian of the EM lower bound is a Lyapunov function of the overall SCVB0 algorithm and the set of fixed points of the EM algorithm.  The boundedness condition, namely that the state variables stay within a compact subset of the state space, follows by observing that $0 < ||\bar{s}_c(x^{(t+1)}, \hat{s})||_1 \leq C$ for every EM statistic $c$, so if the initial state also satisfies this, by convexity $ \hat{\bar{s}}$ will always have its L1 norm similarly bounded.  Having demonstrated that the assumptions required by Theorem 2.3 of Andreiu et al. hold, the convergence result follows. 

\section{Experiments}

In this section we describe an experimental analysis of the proposed SCVB0 algorithm with comparison to the stochastic variational Bayes algorithm of Hoffman et al., hereafter referred to as SVB.  As well as performing an analysis on several large-scale problems, we also investigate the effectiveness of the stochastic LDA inference algorithms at learning topics in near real-time on small corpora.

\subsection{Large-Scale Experiments}

We studied the performance of the algorithms on three large corpora.  The corpora are:

\begin{itemize}
\item \emph{PubMed Central}: A corpus of full-text scientific articles from the open-access PubMed Central database of scientific literature in the biomedical and life sciences.  After processing to remove rare words and stopwords, the corpus contained approximated 320M tokens across 165,000 articles, with a vocabulary size of around 38,500 words.
\item \emph{New York Times}: A corpus containing 1.8 million articles from the New York Times, published between 1987 and 2007.  After processing, the corpus had a dictionary of about 50,000 words and contained 475M distinct tokens.
\item \emph{Wikipedia}: This collection contains 4.6 million articles from the online encyclopedia Wikipedia.  We used the dictionary of 7,700 words extracted by Hoffman et al. for their experiments on an earlier extracted Wikipedia corpus.  There were 811M tokens in the corpus.
\end{itemize}

We explored predictive performance versus wall-clock time between SCVB0 and SVB.  To compare the algorithms fairly, we implemented both of them in the fast high-level language Julia \cite{bezanson2012julia}.  Our implementation of SVB closely follows the python implementation provided by Hoffman, and has several optimizations not mentioned in the original paper including handling the latent topic assignments $z$ implicitly, ``clumping'' of like tokens and sparse updates of the topic matrix.  Our algorithm was implemented as it is written in Algorithm \ref{alg:scvb0}, using the clumping optimization but with no additional algorithmic optimizations.  Specifically, neither implementation used the complicated optimizations taking advantage of sparsity that are exploited by the Vowpal Wabbit implementation of SVB\footnote{\url{https://github.com/JohnLangford/vowpal_wabbit/wiki}} and in the variant of SVB proposed by Mimno \cite{mimno2012sparse}, but instead represent a ``best-effort'' attempt to implement each algorithm efficiently yet following the spirit of the original pseudo-code.

In all experiments, each algorithm was trained using minibatches of size 100.  We used a step-size schedule of $\frac{s}{(\tau + t)^\kappa}$ for document iteration $t$, with $s = 10$, $\tau = 1000$ and $\kappa = 0.9$.  For SCVB0, the document parameters were updated using the same schedule with $s$ = 1, $\tau = 10$ and $\kappa = 0.9$.  We used LDA hyper-parameters $\alpha = 0.1$ and $\eta = 0.01$ for SCVB0.  For SVB, we tried both these same hyperparameter values as well as shifting by $0.5$ as recommended by \cite{asuncionsmoothing} to compensate for the implicit bias in how uncollapsed VB treats hyper-parameters.  We used a single pass to learn document parameters for SCVB0, and tried both a single pass and five passes for SVB.

\begin{figure}
\includegraphics[width=0.7\columnwidth]{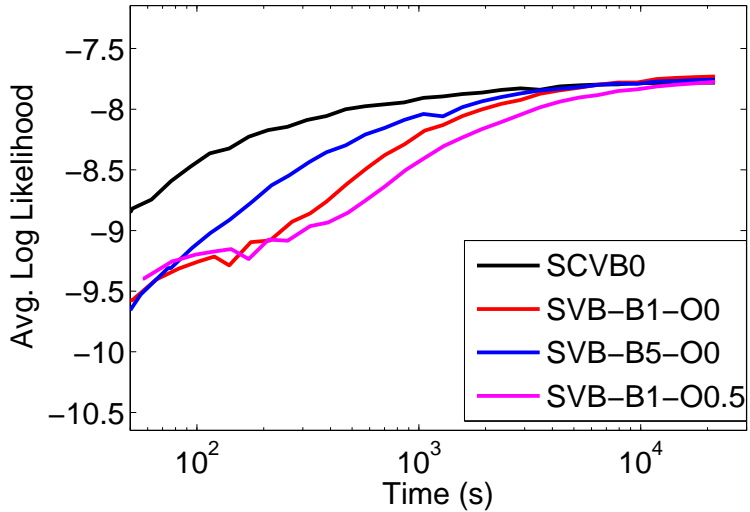}
\caption{Log-likelihood vs Time for the PubMed Central experiments.  SVB-B$x$-O$y$ corresponds to running SVB with $x$ burn-in passes and with hyper-parameters offset from $\alpha = 0.1$ and $\eta=0.01$ by $y$.}
\label{fig:pmc_ll}
\end{figure}

\begin{figure}
\includegraphics[width=0.7\columnwidth]{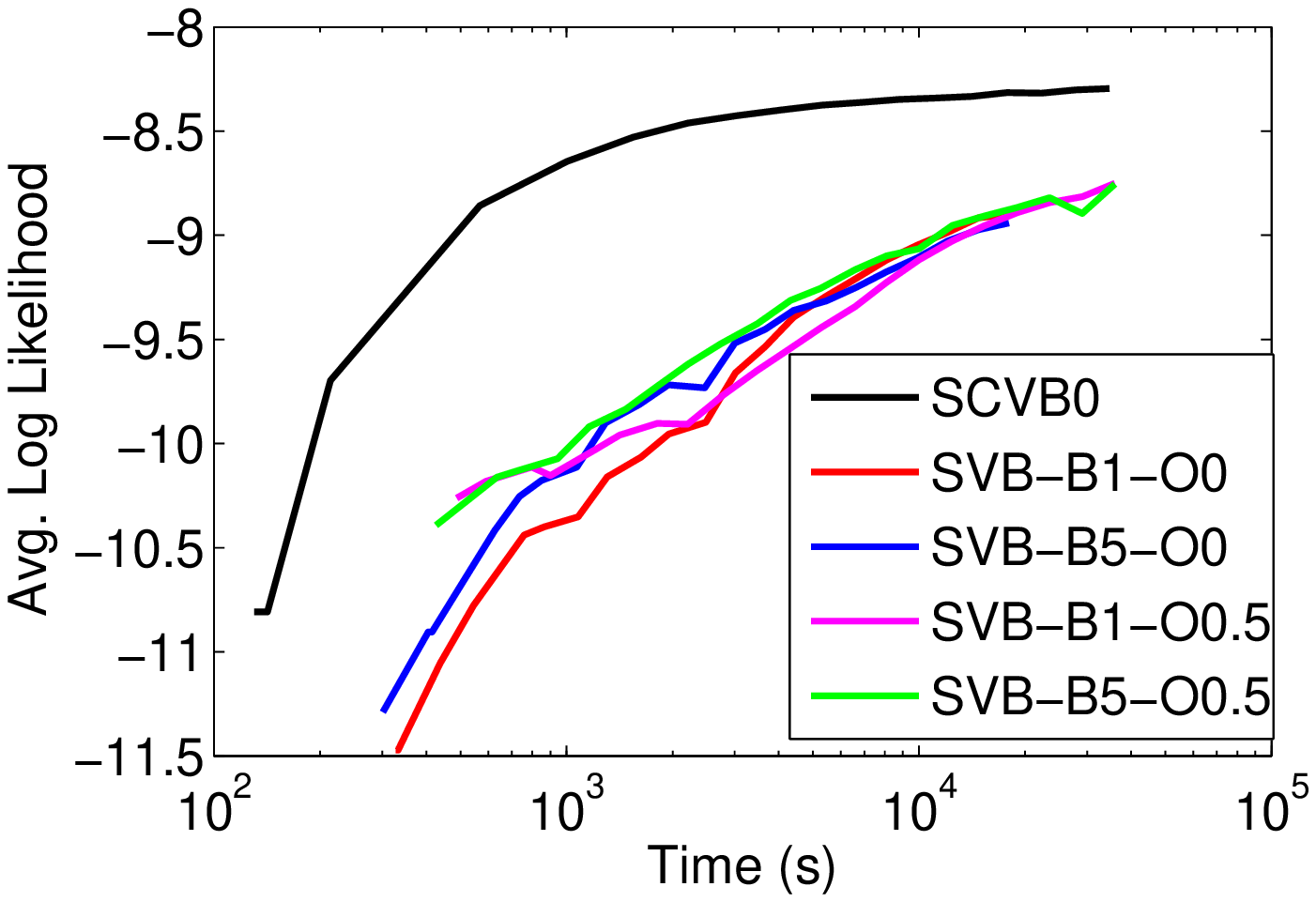}
\caption{Log-likelihood vs Time for the New York Times experiments.  SVB-B$x$-O$y$ corresponds to running SVB with $x$ burn-in passes and with hyper-parameters offset from $\alpha = 0.1$ and $\eta=0.01$ by $y$.}
\label{fig:nyt_ll}
\end{figure}

\begin{figure}
\includegraphics[width=0.7\columnwidth]{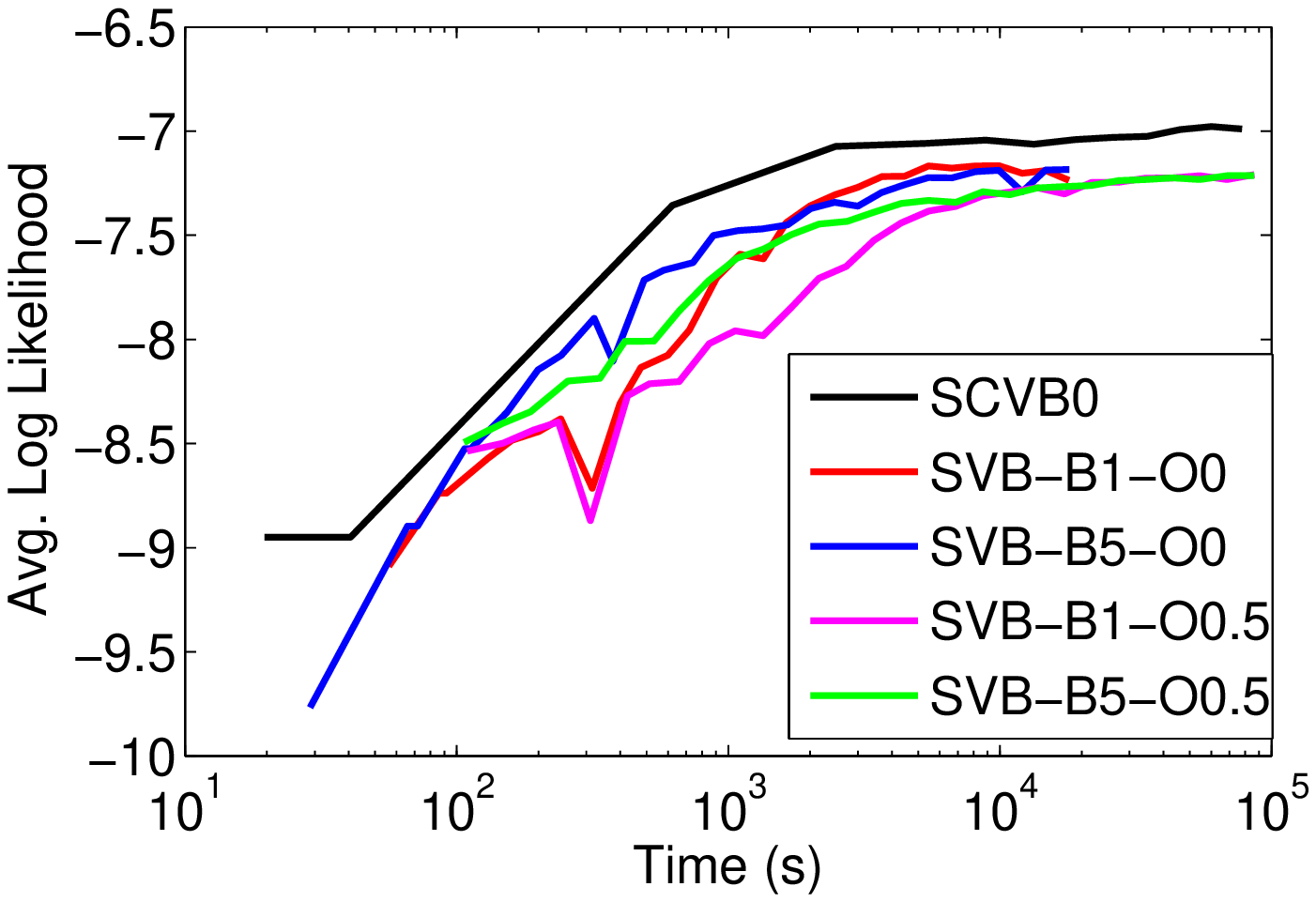}
\caption{Log-likelihood vs Time for the Wikipedia experiments.  SVB-B$x$-O$y$ corresponds to running SVB with $x$ burn-in passes and with hyper-parameters offset from $\alpha = 0.1$ and $\eta=0.01$ by $y$.}
\label{fig:wiki_ll}
\end{figure}

For each experiment we held out 10,000 documents and trained on the remaining documents.  We split each test document in half, estimated document parameters on one half and computed the log-probability of the remaining half of the document.  Figures \ref{fig:pmc_ll} through \ref{fig:wiki_ll} show held-out log-likelihood versus wall-clock time for each algorithm.

\begin{figure}
\includegraphics[width=0.7\columnwidth]{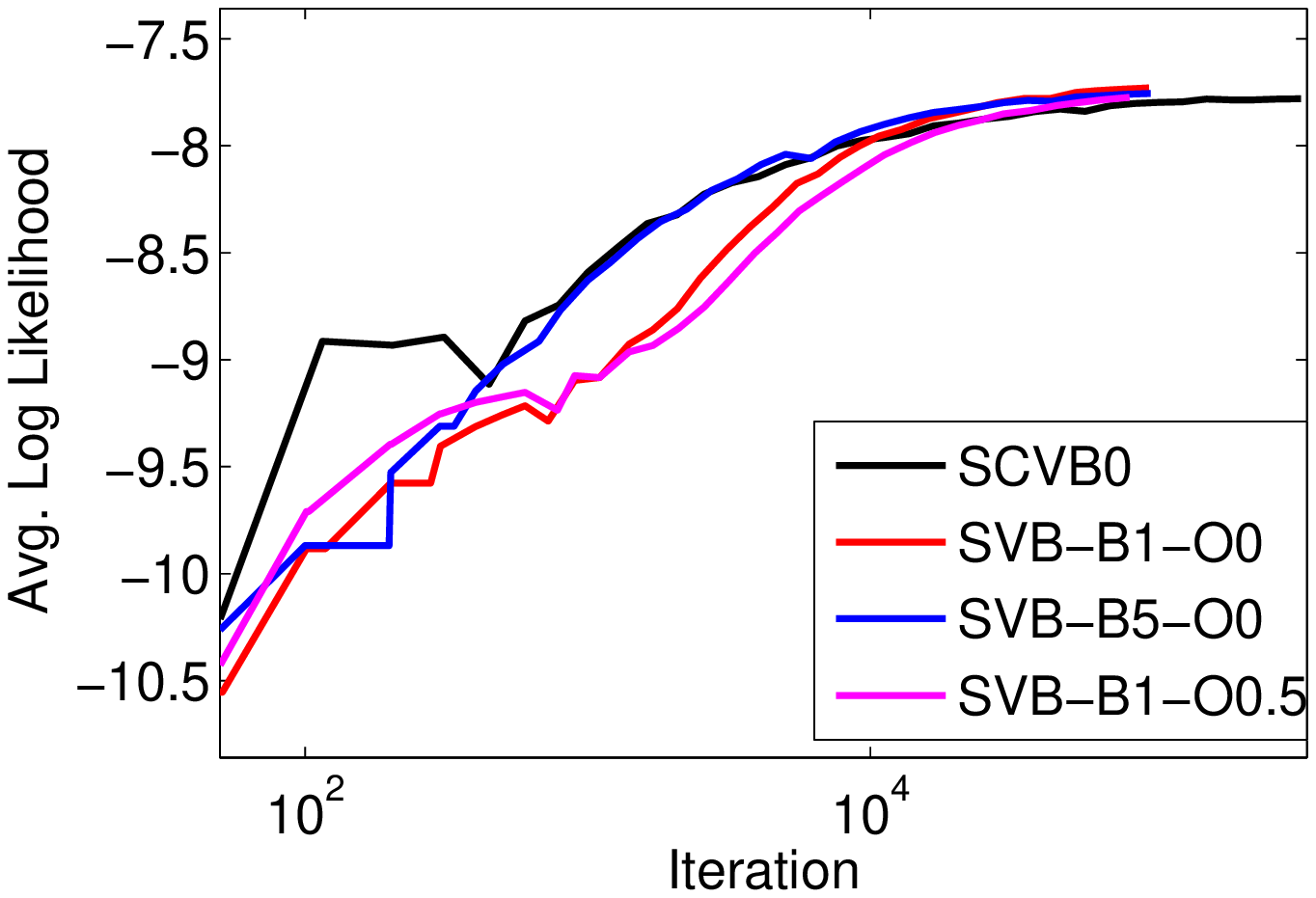}
\caption{Log-likelihood vs Iteration for the PubMed Central experiments.  SVB-B$x$-O$y$ corresponds to running SVB with $x$ burn-in passes and with hyper-parameters offset from $\alpha = 0.1$ and $\eta=0.01$ by $y$.}
\label{fig:pmc_ll_iter}
\end{figure}

For the PubMed Central data, we found that all algorithms perform similarly after about an hour, but prior to that SCVB0 is better, indicating that SCVB0 makes better use of its time.  All algorithms perform similarly per-iteration (see Figure \ref{fig:pmc_ll_iter}), but SCVB0 is able to benefit by processing more documents in the same amount of time.  The per-iteration plots for the other datasets were similar.

Our experiments show that SCVB0 shows a more substantial benefit when employed on larger datasets.  In both the New York Times and Wikipedia experiments SCVB0 converged to a better solution than SVB for any of its parameter settings.  Furthermore, SCVB0 outperforms SVB throughout the run.

\subsection{Small-Scale Experiments}

Stochastic algorithms for LDA have previously only been used on large corpora, however they have the potential to be useful for finding topics very quickly on small corpora as well.  The ability to learn interpretable topics in a matter of seconds is very beneficial for exploratory data analysis (EDA) applications, where a human is in the loop.  Near real-time topic modeling opens the way for the use of topic models in interactive software tools for document analysis.

We investigated the performance of the stochastic algorithms in the small-scale scenario using a corpus of 1740 scientific articles from the machine learning conference NIPS, between 1987 and 1999.  We ran the two stochastic inference algorithms for 5 seconds each, using the parameter settings from the previous experiments but with 20 topics.  Each algorithm was performed ten times.  In the five seconds of training, SCVB0 was typically able to examine 3300 documents, while SVB was typically able to examine around 600 documents.

With the EDA application in mind, we performed a human-subject experiment in the vein of the experiments proposed by Chang and Blei \cite{boyd2009reading}.  The sets of topics returned by each run were randomly assigned across seven human subjects. The participants were all machine learning researchers with technical expertise in the subjects of interest to the NIPS community.  The subjects did not know which algorithms generated which runs.  The top ten words of the topics in each run were shown to the subjects, who were given the following instructions:

\begin{quotation}
  Here are 20 collections of related words.  Some words may not seem to ``belong'' with the other words.  Count the total number of words in each collection that don't ``belong''.
\end{quotation}

This task finds the number of ``errors'' that a topic model inference algorithm makes, relative to human judgement.  It was found that the SCVB0 algorithm had 0.76 errors per topic on average, with a standard deviation of 1.1, while SVB had 1.6 errors per topic on average, with standard deviation 1.2.  A one-sided two sample t-test rejected the hypothesis that the means of the errors per topic were equal, with significance level $\alpha = 0.05$.  Example topics are shown in Table \ref{fig:NIPS}.

We also performed a similar experiment on Amazon Turk involving 52 people using the New York Times corpus. We ran the two stochastic inference algorithms for 60 seconds each using the same parameter settings as above but with 50 topics. Each user was presented with 20 random topics from each algorithm. Example topics are shown in Table \ref{tab:nyt_topics}. Again, the subjects did not know which algorithms generated each set of topics. We included two easy questions with obvious  answers and removed results from users who did not answer them correctly. Comparing the number of ``errors'' for SCVB0 to SVB for each user, we find that SCVB0 had significantly fewer errors for the sampled population at the  $\alpha=.05$ level using a paired t-test, with p-value $< .001$.

\begin{table}
\begin{center}
\begin{tabular}{|c|c|c||c|c|c|}
\hline
\multicolumn{3}{|c||}{SCVB0}& \multicolumn{3}{|c|}{SVB}\\
\hline
 receptor&     data&  learning&  model& results& visual       \\  
 protein&     classification& function& set&   learning&  data    \\   
 secondary& vector&  network&   data&  distribution& activity \\
 proteins&     class&   neural&   training&  information& saliency  \\
 transducer&classifier&  networks&  learning&  map&  noise    \\
 binding&     set&     time&      error& activity&  similarity   \\      
 concentration& algorithm&  order&  parameters&  time&   model   \\ 
 odor&          feature&   error&  markov& figure&  neural     \\     
 morphology & space&  dynamics& estimate& networks&  representations \\
 junction &  vectors&  point&   speech&  state&  functions  \\
\hline
\end{tabular}
\end{center}
\caption{\label{fig:NIPS}Randomly selected example topics after five seconds running time on the NIPS corpus.}
\end{table}

\input{topictable.tex}

\section{Discussion / Related Work}

Connections can be drawn between SCVB0 and other methods in the literature.  The SCVB0 scheme is reminiscent of the online EM algorithm of Cappe and Moulines \cite{cappe2009line}, which also alternates between per data-point parameter updates and online estimates of the expected values of sufficient statistics.  Online EM optimizes the EM lower bound on the log-likelihood in the M-step and computes online averages of exponential family sufficient statistics, while SCVB0 (approximately) updates the mean-field evidence lower bound in the M-step and computes online averages of sufficient statistics required for a CVB0 update in the E-step.

The SCVB0 algorithm also has a very similar structure to SVB, alternating between passes through a document (the optional ``burn-in'' passes) to learn document parameters, and updating variables associated with topics.  However, SCVB0 is stochastic at the word-level while SVB is stochastic at the document level.  In the general framework of Hoffman et al., inference is performed on ``local'' parameters specific to a data point, which are used to perform a stochastic update on the ``global'' parameters. For SVB, the document parameters $\Theta_j$ are local parameters for document $j$, and topics are global parameters.   For SCVB0, the $\gamma_{ij}$'s are local parameters for a word, and both document parameters $N^\Theta$ and topic parameters $N^\Phi$ are global parameters.  This means that updates to document parameters can be made before processing all of the words in the document.

The incremental algorithm of Banerjee and Basu \cite{banerjee2007topic} for MAP inference in LDA is also closely related to the proposed algorithm.  They estimate topic probabilities for each word in sequence, and update MAP estimates of $\Phi$ and $\Theta$ incrementally, using the expected assignments of words to topics in the current document.  SCVB0 can be understood as the collapsed, stochastic variational version of Banerjee and Basu's incremental uncollapsed MAP estimation algorithm.  Interpreting SCVB0 as a MAP estimation algorithm, SCVB0 is the online EM algorithm for MAP estimation operating on the unnormalized representation of LDA, while Banerjee and Basu's algorithm is the incremental EM algorithm operating on the usual normalized representation of LDA.

Another stochastic algorithm for LDA, due to Mimno et al. \cite{mimno2012sparse}, operates in a partially collapsed space, placing it in-between SVB and SCVB0 in terms of representation.  Their algorithm collapses out $\mathbf{\Theta}$ but does not collapse out $\mathbf{\Phi}$.  Estimates of online natural gradient update directions are computed by performing Gibbs sampling on the topic assignments of the words in each document, and averaging over the samples.  The gradient estimate is non-zero only for word-topic pairs which occurred in the samples.  When implemented in a sparse way, the updates scale sub-linearly in the number of topics, causing large improvements in high-dimensional regimes, however these updates are less useful in when the number of topics is smaller (around 100 or less).  The performance gains of this algorithm depend on a careful implementation that takes advantage of the sparsity.  For SCVB0, the minibatch updates are sparse in the rows (words), so some performance enhancements along the lines of those used by Mimno et al. are likely to be possible.

There has been a substantial amount of work on speeding up LDA inference in the literature.   Porteous et al. \cite{porteous2008fast} improved the efficiency of the sampling step for the collapsed Gibbs sampler, and \cite{yao2009efficient} explore a number of alternatives for improving the efficiency of LDA.  The Vowpal Wabbit system for fast machine learning, due to John Langford and collaborators, has a version of SVB that has been engineered to be extremely efficient.  Parallelization is another approach for improving the efficiency of topic models.  Newman et al. \cite{newman2009distributed} introduced an approximate parallel algorithm for LDA where data is distributed across multiple machines, and an exact algorithm for an extension of LDA which takes into account the distributed storage. Smola and Narayanamurthy developed an efficient architecture for parallel LDA inference \cite{smola2010architecture}, using a distributed (key, value) storage for synchronizing the state of the sampler between machines.

\section{Conclusions}

We have introduced SCVB0, an algorithm for performing fast stochastic collapsed variational inference in LDA, and shown that it outperforms stochastic VB on several large document corpora, converging faster and often to a better solution.   The algorithm is relatively simple to implement, with intuitive update rules consisting only of basic arithmetic operations.  We also found that the algorithm was effective at learning good topics from small corpora in seconds, finding topics that were superior than those of stochastic VB according to human judgement.

There are many directions for future work.  The speed of the method could potentially be improved by exploiting sparsity, using techniques such as those employed my Mimno et al. \cite{mimno2012sparse}.  Furthermore, the collapsed representation facilitates the use of the parallelization techniques explored by Newman et al. in \cite{newman2009distributed}.  Finally, SCVB0 could be incorporated into an interactive software tool for exploring the topics of document corpora in real-time.
  

\section{Acknowledgments}
We would like to thank Arthur Asuncion for many helpful discussions.


\bibliographystyle{abbrv}
\bibliography{scvb0}

\appendix
\section{Derivation of the Unnormalized MAP Algorithm}
\label{sec:app_unnormalizedMAP}

Here, we give a more complete derivation of the MAP\_LDA\_U algorithm than is provided in Asuncion et al \cite{asuncionsmoothing}, and show that using a certain ordering of the EM updates results in an algorithm which is very similar to CVB0.

MAP estimation aims to maximize the log posterior probability of the parameters,

\begin{align}
\nonumber
\log Pr(\Theta, \Phi|w, \eta, \alpha) &= \sum_j \log{Pr(w_j| \Theta_j, \Phi )} \\
&+ \sum_{jk} (\alpha - 1) \log(\theta_{jk}) + \sum_{wk} (\eta - 1) \log(\phi_{wk}) + \mbox{ const} \mbox{.}
\end{align}

This objective function cannot readily be optimized directly via, e.g., a gradient update, since the log-likelihood term and its gradient require an intractable sum over $z$ inside the logarithm.  Instead, EM may be performed.  A standard Jensen's inequality argument gives the EM objective function as described by Neal and Hinton \cite{neal1998view}, which, when applied to the MAP estimation problem, is a lower bound $\mathcal{L}(\Theta, \Phi, \bar{\gamma})$ on the posterior probability (CF Bishop \cite{bishop2006pattern}),

\begin{equation}
  \log Pr(\Theta, \Phi|X) \geq \mathcal{L}(\Theta, \Phi, \bar{\gamma}) \triangleq R(\Theta, \Phi, \bar{\gamma}) - \sum_{ijk}\bar{\gamma}_{ijk} \log \bar{\gamma}_{ijk} \mbox{ ,}
\end{equation}
where 

\begin{align}
\nonumber
 R(\Theta, \Phi; \Theta^{(t)}, \Phi^{(t)}) &= \sum_{wk}{(\sum_{ij:w_{ij}=w}\bar{\gamma}_{ijk} + \eta - 1)}\log\phi_{wk}\\
 &+ \sum_{jk}{(\sum_i \bar{\gamma}_{ijk} + \alpha - 1)}\log \theta_{jk} + \mbox{const} 
\end{align}

is the expected complete data log-likelihood, plus terms arising from the prior, and the $\bar{\gamma}_{ijk}$'s are E-step ``responsibilities'',

\begin{equation}
\label{eqn:em_e_step}
\bar{\gamma}_{ijk} \triangleq Pr(z_{ij} = k| \Theta, \Phi, w_{ij}) \propto Pr(w_{ij} | z_{ij} = k, \Theta, \Phi) Pr (z_{ij} = k | \Theta, \Phi) = \phi_{w_{ij}k} \theta_{jk} \mbox{ .}
\end{equation}

Adding Lagrange terms $-\sum_k{\lambda^\Phi_k(\sum_w \phi_{wk} - 1)}$ and $-\sum_j{\lambda^\Theta_j(\sum_k \theta_{jk} - 1)}$, taking derivatives and setting to zero, we obtain the following M-step updates:

\begin{align}
\phi_{wk} &:\propto \sum_{ij: w_{ij} = w}\bar{\gamma}_{ijk} + \eta - 1 \label{eqn:em_phi}&  \theta_{jk} &:\propto \sum_i \bar{\gamma}_{ijk} + \alpha - 1  \mbox{ .} 
\end{align}

It is possible to reparameterize the EM algorithm for LDA in terms of unnormalized counts of the EM ``responsibilities'' instead of $\Theta$ and $\Phi$ \cite{asuncionsmoothing}, which we refer to as the \emph{EM statistics}.  Their definitions are given in Equation \ref{eqn:emstat1}.

Substituting these values into the M-step updates above, then substituting the optimal (M-step updated) parameter assignments into the EM bound and rearranging, we obtain a reparameterization of the EM bound

\begin{align}
\label{eqn:unnormalizedEM} \nonumber
 \log Pr(\Theta, \Phi|X) \geq \sum_{wk}&{(\sum_{ij:w_{ij}=w}\bar{\gamma}_{ijk} + \eta - 1)}\log(\hat{\bar{N}}^{\Phi}_{wk} + \eta - 1)\\ \nonumber
 &+ \sum_{jk}{(\sum_i \bar{\gamma}_{ijk} + \alpha - 1)}\log(\hat{\bar{N}}^{\Theta}_{jk} + \alpha - 1)\\ \nonumber
&-\sum_{k}{(\sum_{ij}\bar{\gamma}_{ijk} + W(\eta - 1))}\log(\hat{\bar{N}}^Z_k + W (\eta - 1))\\
 &- \sum_{ijk}\bar{\gamma}_{ijk} \log \bar{\gamma}_{ijk} + \mbox{const}
\end{align}

where $\mathbf{\hat{\bar{N}}^{\Phi}}$, $\mathbf{\hat{\bar{N}}^{\Theta}}$ and $\mathbf{\hat{\bar{N}}^{Z}}$ are current estimates of the EM statistics, not necessarily synchronized with the $\bar{\gamma}$'s.  To derive M-step updates for this reparameterized formulation, we first add Lagrangian terms to enforce the constraints that each of the EM statistics sums to the number of words in the corpus $C$, $-\lambda_\Phi (\sum_{wk}\hat{\bar{N}}^{\Phi}_{w k} - C)$, $-\lambda_\Phi (\sum_{jk}\hat{\bar{N}}^{\Theta}_{jk} - C)$, $\lambda_Z (\sum_{k}\hat{\bar{N}}^{Z}_{ k} - C)$.  In the following, we derive the update for $\hat{\bar{N}}^\Phi_{wk}$; the derivation is similar for the other parameters.  We take derivatives with respect to each parameter and set them to zero, and plug the constraint equations back into the resulting equations:

\begin{align}
\label{eqn:opt_N_wk}
\frac{\sum_{ij:w_{ij}=w}\bar{\gamma}_{ijk} + \eta - 1}{\hat{\bar{N}}^{\Phi}_{wk} + \eta - 1} -\lambda_{\Phi} &= 0\\ \nonumber
\sum_{ij:w_{ij}=w}\bar{\gamma}_{ijk} + \eta - 1 &= \lambda_{\Phi} (\hat{\bar{N}}^{\Phi}_{wk} + \eta - 1)\\ \nonumber
\hat{\bar{N}}^{\Phi}_{wk} &= \frac{\sum_{ij:w_{ij}=w}\bar{\gamma}_{ijk} + \eta - 1}{\lambda_{\Phi}} - (\eta - 1)\\ \nonumber
C = \sum_{wk}\hat{\bar{N}}^{\Phi}_{wk} &= \frac{1}{\lambda_{\Phi}}\sum_{wk}\big(\sum_{ij:w_{ij}=w}\bar{\gamma}_{ijk} + \eta - 1\big ) - KW(\eta - 1) \mbox{ .}\nonumber
\end{align}

  Solving for the Lagrange multipliers, they turn out to be one:
\begin{align}
\lambda_{\Phi} = \frac{\sum_{wk}{\sum_{ij:w_{ij}=w}\bar{\gamma}_{ijk}} + KW(\eta - 1)}{C + KW(\eta - 1)} = \frac{C + KW(\eta - 1)}{C + KW(\eta - 1)} = 1 \mbox{ .} \nonumber
\end{align}  
  
Plugging this back into Equation \ref{eqn:opt_N_wk} (in the case of $\hat{\bar{N}}^{\Phi}_{wk}$), we obtain M-step updates which synchronize the EM statistics with their definitions in Equation \ref{eqn:emstat1} (i.e. the update in Equation \ref{eqn:unnormalized_map_m_step}).  Note that after the M-step, $\sum_w \hat{\bar{N}}^{\Phi}_{wk} = \hat{\bar{N}}^Z_k \mbox{ } \forall k$, and we did not need to enforce this explicitly in the algorithm.

The E-step finds the expected value of the complete-data log-likelihood, as encoded by the responsibilities $\bar{\gamma}_{ij}$.  Plugging in the estimates of $\mathbf{\Theta}$ and $\mathbf{\Phi}$ from Equation \ref{eqn:recoverMAP} into Equation \ref{eqn:em_e_step} gives us the update in Equation \ref{eqn:map_lda}.  Alternatively, adding Lagrange terms $\sum_{ij}\lambda_{ij}(\sum_k \bar{\gamma}_{ijk} - 1)$ to the bound to enforce the constraint that the $\bar{\gamma}$'s sum to one, setting the derivatives to zero then solving for $\bar{\gamma}_{ij}$ also gives us Equation \ref{eqn:map_lda}.

The standard EM algorithm alternates between complete E and M-steps, i.e. updating all of the $\bar{\gamma}_{ij}$'s, followed by synchronizing the EM statistics with the responsibilities.  However, the EM algorithm can be viewed as a coordinate ascent algorithm on the lower bound objective function, and partial E and M-steps also improve this bound \cite{neal1998view}.  In our case, both updating a single $\bar{\gamma}_{ij}$, and subsequently synchronizing the EM statistics to reflect the new value (partial E and M-steps, respectively) are coordinate ascent updates which improve the EM lower bound in Equation \ref{eqn:unnormalizedEM}.  So an algorithm that iteratively performs the update in Equation \ref{eqn:map_lda} for each token (a partial E-step), while continuously keeping the EM statistics in synch with the $\bar{\gamma}_{ij}$'s as in Equation \ref{eqn:emstat1} (a partial M-step), is equivalent to the above EM algorithm but merely performing the coordinate ascent updates in a different order.  This algorithm is very similar to CVB0, but using Equation \ref{eqn:map_lda} instead of Equation \ref{eqn:cvb0}.

\section{Lyapunov Function}
\label{sec:app_lyapunov}

A Lyapunov function is a function which gives a stochastic analogue of the monotonicity property of the EM algorithm, the existence of which is a standard argument for the stability and convergence of a stochastic approximation algorithm.  Theorem 2.3 of Andreiu et al. \cite{andrieu2005stability} states that under mild conditions, convergence is assured for a Robins-Monro SA algorithm endowed with a Lyapunov function with certain properties.  Andreiu et al. consider an SA with state space $\mathbf{\Theta}$ for finding $h(\theta) = \mathbf{0}$, where $\mathbf{\Theta}$ is an open subset of $\mathbb{R}^n$, and $h: \mathbf{\Theta} \rightarrow \mathbb{R}^n$.  They require the existence of a continuously differentiable function $w:  \mathbf{\Theta} \rightarrow [0, \infty)$, the Lyapunov function, such that the following conditions hold:
\begin{itemize}
\item (i) There exists $M_0 > 0$ such that
$$\mathcal{L} \triangleq \{\theta \in \mathbf{\Theta}, \langle\nabla w(\theta), h(\theta) \rangle = 0\} \subset \{\theta \in \mathbf{\Theta}, w(\theta) < M_0 \}$$,
\item (ii) There exists $M_1 \in (M_0, \infty]$ such that $\{\theta \in \mathbf{\Theta}, w(\theta) < M_1\}$ is a compact set,
\item (iii) For any $\theta \in \mathbf{\Theta} \ \mathcal{L}, \langle \nabla w(\theta), h(\theta) \rangle < 0$
\item (iv) $w(\mathcal{L})$ has an empty interior.
\end{itemize}

In our case, recall that in Section \ref{sec:convergence} we showed that the SCVB0 updates for each of the EM statistics $c$ corresponds to a Robbins-Monro SA for finding the zeros of $f_c(X, \hat{s}^{(t)}) -\hat{s}^{(t)}_c$, i.e. the fixed points of MAP\_LDA\_U for $\hat{s}_c$.  In the overall algorithm, $\theta = (\hat{\bar{N}}^\Theta, \hat{\bar{N}}^\Phi, \hat{\bar{N}}^Z)$, and $h(\theta)$ is the direction of the M-step update that we would take if we were to first perform a full E-step.  Finding $h(\theta) = \mathbf{0}$, as the SA algorithm is designed to do, corresponds to finding the fixed points of the MAP\_LDA\_U EM algorithm, which are at the stationary points of the posterior distribution of the parameters, i.e. the objective function for MAP estimation.

We will now show that minus the EM lower bound, augmented with Lagrange terms, is a Lyapunov function of the overall algorithm.  As we found in Appendix \ref{sec:app_unnormalizedMAP}, if we include Lagrange constraints in the EM bound to ensure that the EM statistics sum to $C$, set the gradient to zero and solve for the Lagrange multipliers, the Lagrange multipliers turn out to equal one.  Substituting this value into the Lagrangian and dropping constant terms, we have our candidate function

\begin{align}
 \nonumber
-w(\hat{\bar{N}}^\Theta, \hat{\bar{N}}^\Phi, \hat{\bar{N}}^Z) \triangleq \sum_{wk}&\Big [ {(\sum_{ij:w_{ij}=w}\bar{\gamma}_{ijk} + \eta - 1)}\log(\hat{\bar{N}}^{\Phi}_{wk} + \eta - 1) -\hat{\bar{N}}^{\Phi}_{wk} \Big]\\ \nonumber
 &+ \sum_{jk}\Big [{(\sum_i \bar{\gamma}_{ijk} + \alpha - 1)}\log(\hat{\bar{N}}^{\Theta}_{jk} + \alpha - 1) -\hat{\bar{N}}^{\Theta}_{jk} \Big ]\\ \nonumber
&-\sum_{k} \Big [ {(\sum_{ij}\bar{\gamma}_{ijk} + W(\eta - 1))}\log(\hat{\bar{N}}^Z_k + W (\eta - 1)) -\hat{\bar{N}}^Z_k \Big]\\
 &- \sum_{ijk}\bar{\gamma}_{ijk} \log \bar{\gamma}_{ijk} \mbox{ ,}
\end{align}

where $\bar{\gamma}$ are E-step estimates computed from the current EM statistics -- note that $w(\theta)$ is not a function of them.  We want to show that conditions (i) -- through (iv) hold for $w(\theta)$.

Condition (ii) holds because the EM statistics have L1 norm bounded by $C$.  Condition (iv) holds by Sard's theorem.  The key conditions are (i) and (iii), which involve the directional derivative of $w(\theta)$ at $\theta$ along $h(\theta)$, $\langle\nabla w(\theta), h(\theta) \rangle$ (where we have appended the EM statistics so that $\theta$ is a vector).  This is the instantaneous change in $w(\theta)$ in the direction of the EM update.  Note that a step with a step-size multiplier of one in the direction $h(\theta)$ is guaranteed by the monotonicity of EM to improve the (Lagrangian of the) lower bound, and thereby lower $w(\theta)$.  However, we have to check that an infintesimal step in that direction also improves this function.

Suppose we are not at a fixed point of EM, i.e. $h(\theta) \neq \mathbf{0}$.  Fixing $\bar{\gamma}$ to E-step-updated values based on $\theta$, we know from the derivation of the M-step update that the Lagrangian of the EM lower bound has a unique maximum at the M-step-updated value, located at $\theta + h(\theta)$.  Since this maximum is unique and there are no other stationary points, each point in the direction  $h(\theta)$ of the maximum has an increasingly large value of the Lagrangian of the EM bound, holding $\bar{\gamma}$ fixed.  These values computed with $\bar{\gamma}$ fixed to its current value are a lower bound on the Lagrangian $-w(\theta)$ at those points, as $w(\theta)$ is computed using E-step updated $\bar{\gamma}$'s which must strictly improve the EM lower bound relative to the current (or any other) $\bar{\gamma}$.  So every point on the line segment between $\theta$ and $\theta + h(\theta)$ has a strictly higher value of the Lagrangian $-w(\theta)$ than at $\theta$, i.e. $-w(\theta + \lambda h(\theta)) - (-w(\theta)) > 0 \mbox{, } \forall \lambda \in (0,1]$.  This implies that $\langle\nabla w(\theta), h(\theta) \rangle = \lim_{\lambda \rightarrow 0} \frac{w(\theta + \lambda h(\theta)) - w(\theta)}{\lambda} < 0$, and (iii) holds.\footnote{The directional derivative of $w$ at $\theta$ along $v$ is defined to be $\lim_{\lambda \rightarrow 0} \frac{w(\theta + \lambda v) - w(\theta)}{\lambda}$.  If $w$ is differentiable at $\theta$, the directional derivative equals $\langle\nabla w(\theta), v \rangle$.}

At at a fixed point of MAP\_LDA\_U, which (due to the properties of EM) happens IFF the algorithm is at a stationary point of the MAP objective function, $h(\theta) = \mathbf{0}$, and it can be shown by inspection that $\nabla w(\theta) = \mathbf{0}$ only under these conditions also.  In this case, the directional derivative $\langle\nabla w(\theta), h(\theta) \rangle = 0$, and $\theta \in \mathcal{L}$.  So $\theta \in \mathcal{L}$ IFF $\theta$ is at a stationary point of the MAP objective function, and (i) holds.  Along with a boundedness condition demonstrated in Section \ref{sec:convergence}, Theorem 2.3 of Andreiu et al. \cite{andrieu2005stability} now gives us that with an appropriate sequence of step sizes, in the limit as the number of iterations approaches infinity the distance from $\mathcal{L}$ is zero.

\end{document}

%% file: topictable.tex
\begin{table*}[ht]
\begin{center}
\begin{tabular}{|l|l|l||l|l|l|}
  \hline
\multicolumn{3}{|c||}{SCVB} & \multicolumn{3}{c|}{SVB}\\
\hline
county & station & league & president & year & mr \\ 
  district & company & goals & midshipmen & cantatas & company \\ 
  village & railway & years & open & edward & mep \\ 
  north & business & club & forrester & computing & husbands \\ 
  river & services & clubs & archives & main & net \\ 
  area & market & season & iraq & years & state \\ 
  east & line & played & left & area & builder \\ 
  town & industry & cup & back & withdraw & offense \\ 
  lake & stations & career & times & households & obscure \\ 
  west & owned & team & saving & brain & advocacy \\ 
   \hline
\end{tabular}
\end{center}
\caption{\label{tab:nyt_topics}Randomly selected example topics after sixty seconds running time on the NYT corpus.}
\end{table*}

%% file: SCVB0_arXiv.bbl
\begin{thebibliography}{10}

\bibitem{andrieu2005stability}
C.~Andrieu, {\'E}.~Moulines, and P.~Priouret.
\newblock Stability of stochastic approximation under verifiable conditions.
\newblock {\em SIAM Journal on control and optimization}, 44(1):283--312, 2005.

\bibitem{asuncionsmoothing}
A.~Asuncion, M.~Welling, P.~Smyth, and Y.~Teh.
\newblock On smoothing and inference for topic models.
\newblock In {\em Uncertainty in Artificial Intelligence}, 2009.

\bibitem{atkins2012family}
D.~C. Atkins, T.~N. Rubin, M.~Steyvers, M.~A. Doeden, B.~R. Baucom, and
  A.~Christensen.
\newblock Topic models: A novel method for modeling couple and family text
  data.
\newblock {\em Journal of Family Psychology}, 6:816--827, 2012.

\bibitem{banerjee2007topic}
A.~Banerjee and S.~Basu.
\newblock Topic models over text streams: A study of batch and online
  unsupervised learning.
\newblock In {\em SIAM Data Mining}, 2007.

\bibitem{bezanson2012julia}
J.~Bezanson, S.~Karpinski, V.~B. Shah, and A.~Edelman.
\newblock Julia: A fast dynamic language for technical computing.
\newblock {\em CoRR}, abs/1209.5145, 2012.

\bibitem{bishop2006pattern}
C.~M. Bishop et~al.
\newblock {\em Pattern recognition and machine learning}, volume~1.
\newblock springer New York, 2006.

\bibitem{blei2003latent}
D.~Blei, A.~Ng, and M.~Jordan.
\newblock Latent {D}irichlet allocation.
\newblock {\em The Journal of Machine Learning Research}, 3:993--1022, 2003.

\bibitem{boyd2009reading}
J.~Boyd-Graber, J.~Chang, S.~Gerrish, C.~Wang, and D.~Blei.
\newblock Reading tea leaves: How humans interpret topic models.
\newblock In {\em Proceedings of the 23rd Annual Conference on Neural
  Information Processing Systems}, 2009.

\bibitem{cappe2009line}
O.~Capp{\'e} and E.~Moulines.
\newblock On-line expectation--maximization algorithm for latent data models.
\newblock {\em Journal of the Royal Statistical Society: Series B (Statistical
  Methodology)}, 71(3):593--613, 2009.

\bibitem{carpenter2010collapsed}
B.~Carpenter.
\newblock Integrating out multinomial parameters in latent {D}irichlet
  allocation and naive bayes for collapsed {G}ibbs sampling.
\newblock Technical report, LingPipe, 2010.

\bibitem{griffithsSteyvers2004}
T.~Griffiths and M.~Steyvers.
\newblock Finding scientific topics.
\newblock {\em Proceedings of the National Academy of Sciences of the United
  States of America}, 101(Suppl 1):5228, 2004.

\bibitem{hoffman2012stochastic}
M.~Hoffman, D.~Blei, C.~Wang, and J.~Paisley.
\newblock Stochastic variational inference.
\newblock {\em arXiv preprint arXiv:1206.7051}, 2012.

\bibitem{hoffman2010online}
M.~D. Hoffman, D.~M. Blei, and F.~Bach.
\newblock Online learning for latent {D}irichlet allocation.
\newblock {\em Advances in Neural Information Processing Systems}, 23:856--864,
  2010.

\bibitem{mimno2012computational}
D.~Mimno.
\newblock Computational historiography: Data mining in a century of classics
  journals.
\newblock {\em Journal on Computing and Cultural Heritage (JOCCH)}, 5(1):3,
  2012.

\bibitem{mimno2012reconstructing}
D.~Mimno.
\newblock Reconstructing pompeian households.
\newblock {\em Uncertainty in Artificial Intelligence}, 2012.

\bibitem{mimno2012sparse}
D.~Mimno, M.~Hoffman, and D.~Blei.
\newblock Sparse stochastic inference for latent {D}irichlet allocation.
\newblock In {\em Proceedings of the International Conference on Machine
  Learning}, 2012.

\bibitem{minka2004power}
T.~Minka.
\newblock Power {EP}.
\newblock Technical report, Microsoft Research, Cambridge, UK, 2004.

\bibitem{neal1998view}
R.~M. Neal and G.~E. Hinton.
\newblock A view of the em algorithm that justifies incremental, sparse, and
  other variants.
\newblock In M.~Jordan, editor, {\em Learning in graphical models}, pages
  355--368. Springer, 1998.

\bibitem{newman2009distributed}
D.~Newman, A.~Asuncion, P.~Smyth, and M.~Welling.
\newblock Distributed algorithms for topic models.
\newblock {\em The Journal of Machine Learning Research}, 10:1801--1828, 2009.

\bibitem{porteous2008fast}
I.~Porteous, D.~Newman, A.~Ihler, A.~Asuncion, P.~Smyth, and M.~Welling.
\newblock Fast collapsed gibbs sampling for latent dirichlet allocation.
\newblock In {\em Proceedings of the ACM SIGKDD International Conference on
  Knowledge Discovery and Data Mining}, pages 569--577, 2008.

\bibitem{robbins1951stochastic}
H.~Robbins and S.~Monro.
\newblock A stochastic approximation method.
\newblock {\em The Annals of Mathematical Statistics}, pages 400--407, 1951.

\bibitem{sato2012rethinking}
I.~Sato and H.~Nakagawa.
\newblock Rethinking collapsed variational {B}ayes inference for {LDA}.
\newblock {\em Proceedings of the International Conference on Machine
  Learning}, 2012.

\bibitem{smola2010architecture}
A.~Smola and S.~Narayanamurthy.
\newblock An architecture for parallel topic models.
\newblock {\em Proceedings of the VLDB Endowment}, 3(1-2):703--710, 2010.

\bibitem{teh2007collapsed}
Y.~Teh, D.~Newman, and M.~Welling.
\newblock A collapsed variational bayesian inference algorithm for latent
  {D}irichlet allocation.
\newblock {\em Advances in Neural Information Processing Systems}, 19:1353,
  2007.

\bibitem{yao2009efficient}
L.~Yao, D.~Mimno, and A.~McCallum.
\newblock Efficient methods for topic model inference on streaming document
  collections.
\newblock In {\em Proceedings of the 15th ACM SIGKDD international conference
  on Knowledge discovery and data mining}, pages 937--946. ACM, 2009.

\end{thebibliography}
